\documentclass[runningheads]{accv22/llncs}

\usepackage{graphicx}
\usepackage{amsmath,amssymb}
\usepackage{color}
\usepackage{multirow}
\usepackage{booktabs}
\usepackage{caption}
\usepackage{subcaption}
\usepackage{tikz}
\usepackage{pdfpages}
\usepackage{hyperref}

\def\checkmark{\tikz\fill[scale=0.4](0,.35) -- (.25,0) -- (1,.7) -- (.25,.15) -- cycle;} 

\setbox0\hbox{\tabular{@{}l}Ker.\endtabular}

\setbox1\hbox{\tabular{@{}l}Flow\endtabular} 
\setbox2\hbox{\tabular{@{}l}Att.\endtabular} 
\setbox3\hbox{\tabular{@{}l}K\endtabular}
\setbox4\hbox{\tabular{@{}l}K+F\endtabular}
\setbox5\hbox{\tabular{@{}c}No\\Flow\endtabular}

\begin{document}
\pagestyle{headings}
\mainmatter

\title{Cross-Attention Transformer for Video Interpolation}

\titlerunning{Cross-Attention Transformer for Video Interpolation}
\authorrunning{H. Kim et al.}

\author{
Hannah Halin Kim\orcidID{0000-0003-2588-0190} \and 
Shuzhi Yu\orcidID{0000-0003-2514-381X} \and \\
Shuai Yuan\orcidID{0000-0003-4039-0464} \and
Carlo Tomasi\orcidID{0000-0001-6104-6641}
}
\institute{
Duke University, Durham NC 27708, USA\\
\email{\{hannah,shuzhiyu,shuai,tomasi\}@cs.duke.edu}
}

\maketitle

\begin{abstract}

We propose TAIN (Transformers and Attention for video INterpolation), a residual neural network for video interpolation, which aims to interpolate an intermediate frame given two consecutive image frames around it. We first present a novel vision transformer module, named Cross Similarity (CS), to globally aggregate input image features with similar appearance as those of the predicted interpolated frame. These CS features are then used to refine the interpolated prediction. To account for occlusions in the CS features, we propose an Image Attention (IA) module to allow the network to focus on CS features from one frame over those of the other. TAIN outperforms existing methods that do not require flow estimation and performs comparably to flow-based methods while being computationally efficient in terms of inference time on Vimeo90k, UCF101, and SNU-FILM benchmarks.
\end{abstract}


\section{Introduction}

Video interpolation~\cite{cain,Niklaus_ICCV_2017,softsplat,bmbc,dain,sepconv++,vimeo} aims to generate new frames between consecutive image frames in a given video. This task has many practical applications, ranging from frame rate up-conversion~\cite{fruc} for human perception~\cite{hfr}, video editing for object or color propagation~\cite{cp}, slow motion generation~\cite{superslomo}, and video compression~\cite{vidocompression}.

\setlength{\tabcolsep}{2pt}
\begin{figure}[t]
    \centering
     \begin{subfigure}[b]{0.27\textwidth} 
     \centering
     \begin{tabular}{cccc}
        \toprule
        & Network & Inference Time & PSNR \\ \midrule
        \textcolor{teal}{\multirow{3}{*}{\rotatebox{90}{\usebox5}}} 
        & EDSC & \underline{\textbf{19.37} $\pm$ \textbf{0.06}} &  34.84\\
        & CAIN & 25.21 $\pm$ 0.45 & 34.65\\ \cline{2-4}
        & TAIN (ours) & 32.59 $\pm$ 0.82 & \textbf{35.02} \\ \midrule
        \textcolor{red}{
        \multirow{4}{*}{\rotatebox{90}{\usebox1}}}
        & DAIN & 168.56 $\pm$ 0.33 & 34.71 \\
        & BMBC & 333.24 $\pm$ 6.60 & 35.01 \\
        & ABME & \textbf{116.66} $\pm$ \textbf{1.07} & 36.18 \\
        & VFIformer & 476.06 $\pm$ 11.92 & \underline{\textbf{36.50}} \\ 
        \bottomrule
    \end{tabular}
         \label{fig:it_table}
     \end{subfigure}
     \hfill
     \begin{subfigure}[b]{0.47\textwidth}
         \centering
         \includegraphics[width=0.97\textwidth]{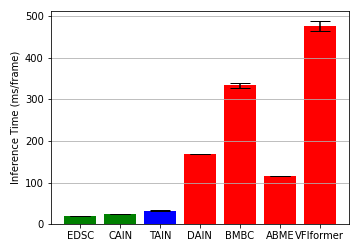} 
         \label{fig:it_figure}
     \end{subfigure}
    \caption{Inference times (in milliseconds per frame prediction) of existing video interpolation methods and TAIN (ours) on a single P100 GPU. The inference times are computed as average and standard deviation over 300 inferences using two random images of size $256\times 256$ as input. As a reference, we also list  performance (PSNR) on Vimeo-90k~\cite{vimeo}. Flow-based methods (red) have much higher inference times compared to TAIN (blue) and other non-flow-based methods (green). Bold values show the best performance in each panel, and underlined bold values show the best performance across both panels. Best viewed in color.}
    \label{fig:inference_time}
\end{figure}

Recent work on video interpolation starts by estimating optical flow in both temporal directions between the input frames. Some systems~\cite{softsplat,bmbc,dain,Niklaus_2018_CVPR,MEMC-Net,Hu_2022_CVPR} use flow predictions output by an off-the-shelf pre-trained estimator such as FlowNet~\cite{flownet} or PWC-Net~\cite{pwc}, while others estimate flow as part of their own pipeline~\cite{vimeo,superslomo,Gui_2020_CVPR,liu2017voxelflow,Xiang_2020_CVPR,Danier_2022_CVPR,Lu_2022_CVPR,abme,Choi_2021_ICCV}. The resulting bi-directional flow vectors are interpolated to infer the flow between the input frames and the intermediate frame to be generated. These inferred flows are then used to warp the input images towards the new one. 

While these flow-based methods achieve promising results, they also come with various issues as follows.
First, they are computationally expensive and rely heavily on the quality of the flow estimates (see Figure \ref{fig:inference_time}). Specifically, state-of-the art flow estimators~\cite{flownet,pwc,raft} require to compute at least two four-dimensional cost volumes at all pixel positions. Second, they are known to suffer at occlusions, where flow is undefined; near motion boundaries, where flow is discontinuous; and in the presence of large motions~\cite{monet,yu}. For instance, RAFT~\cite{raft} achieves an End-Point Error (EPE) of 1.4 pixels on Sintel~\cite{mpi}, but the EPE rises to 6.5 within 5 pixels from a motion boundary and to 4.7 in occlusion regions. Warping input images or features with these flow estimates will lead to poor predictions in these regions. This is especially damaging for video interpolation as motion boundaries and occlusions result from motion~\cite{monet}, and are therefore important regions to consider for motion compensation. Further, these flow estimators are by-and-large trained on synthetic datasets~\cite{mpi,flythings3D} to avoid the cost and difficulties of annotating real video~\cite{Yuan}, and sometimes fail to capture some of the challenges observed in real data, to which video interpolation is typically applied.

Apart from flow inputs, many of these methods also utilize additional inputs and networks to improve performance, which adds to their computational complexity. These include depth maps~\cite{dain}, occlusion maps~\cite{dain,vimeo,superslomo,MEMC-Net,Lee_2020_CVPR}, multiple input frames~\cite{Shi_2022_CVPR} for better flow estimation, image classifiers (\textit{e.g.}, VGG~\cite{vgg}, ResNet~\cite{resnet}) pretrained on ImageNet~\cite{imagenet} for contextual features~\cite{cain,Niklaus_ICCV_2017,softsplat,superslomo,Lee_2020_CVPR}, adversarial networks for realistic estimations~\cite{Lee_2020_CVPR}, and event cameras~\cite{Yang_2015_evencamera,tulyakov2021time,Zhang_2022_CVPR,Tulyakov_2022_CVPR} to detect local brightness changes across frames. Instead, we achieve competitive performance with a single network that uses two consecutive input frames captured with commonly available devices.

The proposed TAIN system (Transformers and Attention for video INterpolation) is a residual neural network that requires no estimation of optical flow. Inspired by the recent success of vision transformers~\cite{dosovitskiy2021image,huang2020ccnet,vaswani2017attention,gma}, we employ a novel transformer-based module named Cross Similarity (CS) in TAIN. Video interpolation requires the network to match corresponding points across frames, and our CS transformer achieves this through cross-attention.
Vision Transformers typically use self-attention to correlate each feature of an image to every other feature \emph{in the same image}. The resulting similarity maps are then used to collect relevant features from other image locations. Our CS module instead compares features \emph{across frames}, namely, between an input frame and the current intermediate frame prediction. High values in this cross-frame similarity map indicate features with similar appearance. These maps are then used to aggregate features from the appropriate input image to yield CS features, and these are used in turn to refine the frame prediction. While there exists recent work on video interpolation that utilize vision transformers~\cite{Lu_2022_CVPR,Shi_2022_CVPR}, our CS is specifically designed for video interpolation as it computes cross-frame similarity rather than within-frame self-similarity.

To account for occlusions or motion boundaries, we use CS scores in an Image Attention (IA) module based on spatial attention~\cite{Zhang_2018_CVPR}. Given a feature in the tentative frame prediction, its maximum similarity score from our CS module will indicate whether or not a similar feature exists in either input frame. This maximum score will be high if such a feature exists and low if it does not. If the feature in the current frame prediction is occluded in one of the input images, the corresponding score will likely be low. Features at positions that straddle a motion boundary in the frame prediction often have a low maximum score as well. This is because boundary features contain information about pixels from both side of the boundary, and the particular mixture of information will change if the two sides move differently. The CS features aggregated from these low maximum similarity scores will not help in refining the current frame prediction, and our IA module learns to suppress them.

Thanks to CS transformer and IA module, TAIN improves or performs comparably to existing methods on various benchmarks, especially compared to those that do not require flow estimation. Our contributions are as follows:
\begin{itemize}
    \item A novel Cross Similarity module based on vision transformer that aggregates features of the input image frames that have similar appearance to predicted-frame features. These aggregated features help refine the frame prediction. 
    \item A novel Image Attention module that gives the predictor the ability to weigh features in one input frame over those in the other. This information is shown to be especially helpful near occlusions and motion boundaries.
    \item State-of-the-art performance on Vimeo-90k, UCF101, and SNU-FILM among methods that do not require optical flow estimation.
\end{itemize}

\section{Related Work}

\subsection{Video Interpolation}
Deep learning has rapidly improved the performance of video interpolation in recent work. Long \textit{et al.}~\cite{long} are the first to use a CNN for video interpolation. In their system, an encoder-decoder network predicts the intermediate frame directly from two image frames using inter-frame correspondences. Subsequent work can be categorized largely into kernel-, flow-, and attention-based.

\textbf{Kernel-based approaches} compute the new frame with convolutions over local patches, and use CNNs to estimate spatially-adaptive convolutional kernels~\cite{Niklaus_ICCV_2017,sepconv++,Niklaus_CVPR_2017}. These methods use large kernel sizes to accommodate large motion, and large amounts of memory are required as a result when frames have high resolution. Niklaus \textit{et. al} propose to use separable convolutional kernels to reduce memory requirements and further improve results. Since kernel-based methods cannot handle motion larger than the pre-defined kernel size, EDSC~\cite{Niklaus_ICCV_2017} estimates not only adaptive kernels, but also offsets, masks, and biases to retrieve information from non-local neighborhoods. EDSC achieves the current state-of-the-art performance among methods that do not require optical flow.
 
\textbf{Flow-based approaches} to video interpolation rely on bi-directional flow estimates using off-the-shelf pre-trained flow estimators (e.g. FlowNetS, PWC-Net)~\cite{softsplat,bmbc,dain,Niklaus_2018_CVPR,MEMC-Net,Hu_2022_CVPR}, or estimate flow as part of their pipeline~\cite{vimeo,superslomo,Gui_2020_CVPR,liu2017voxelflow,Xiang_2020_CVPR,Danier_2022_CVPR,Lu_2022_CVPR,abme,Choi_2021_ICCV}. Most of these methods assume linear motion between frames and use the estimated flow to warp input images and their features to the intermediate time step for prediction. In order to avoid making this linear motion assumption, Gui \textit{et al.}~\cite{Gui_2020_CVPR} do not predict flow between two input frames but instead attempt to produce flow directly between the intermediate frame being predicted and the two input frames. Park \textit{et al.}~\cite{abme} do compute a tentative intermediate frame from the given flow estimates, but they re-estimate flow between the tentative predicted frame and the input images. These new estimates are then used for a final estimation of the intermediate frame. Multiple optical flow maps have also been used to account for complex motion patterns and mitigate the resulting prediction artifacts~\cite{Hu_2022_CVPR,Danier_2022_CVPR}. While these flow-based methods show promising results, they are computationally expensive. We do not require flow estimates in our approach.

Some approaches~\cite{dain,MEMC-Net,Lee_2020_CVPR} integrate \textbf{both flow-based and kernel-based methods} by combining optical flow warping with learned adaptive local kernels. These methods perform robustly in the presence of large motions and are not limited by the assumption of a fixed motion range. They use small kernels that require less memory but are still expensive. 

Recently, work by Choi \textit{et al.}~\cite{cain} proposes a residual network called CAIN that interpolates video through \textbf{attention mechanism}~\cite{Zhang_2018_CVPR} without explicit computation of kernels or optical flow to curb model complexity and computational cost. The main idea behind their design is to distribute the information in a feature map into multiple channels through PixelShuffle, and extract motion information by processing the channels through a Channel Attention module. In our work, we extend CAIN with a novel vision transformer module and spatial attention module, still without requiring flow estimates or adaptive kernels.

\subsection{Vision Transformers}
Transformers~\cite{vaswani2017attention,gma} have shown success in both computer vision~\cite{dosovitskiy2021image,huang2020ccnet,ramachandran2019standalone} and natural language processing~\cite{vaswani2017attention,nlp_att} thanks to their ability to model long-range dependencies. Self-attention~\cite{vaswani2017attention,nlp_att} has shown the most success among various modules in the transformer architecture, and derives query, key, and value vectors from the same image. Due to their content-adaptive nature, transformers have also been applied to video interpolation. Shi~\textit{et. al}~\cite{Shi_2022_CVPR} consider four frame inputs and propose to use a self-attention transformer based on SWIN~\cite{liu2021swin} to capture long-range dependencies across both space and time. Lu~\textit{et. al}~\cite{Lu_2022_CVPR} propose a self-attention transformer along with a flow estimator to model long-range pixel correlation for video interpolation. Their work currently achieves the state-of-the-art performance on various video interpolation benchmarks. 
Different from existing work, we do not use self-attention in our work, but instead use cross-attention. In self-attention, query, key, and value are different projections of the same feature. We use cross-attention, where query and key are the same projections (shared weights) of different features. Specifically, our transformer module selects and refines features based on the similarities between the interpolated frame (query) and the two input frames (keys) while handling the occlusions that occur in the two input frames. On the other hand, existing transformer-based methods compare the features of the input frames without any explicit consideration of the interpolated frame or occlusions.

\setlength{\tabcolsep}{0pt}
\begin{figure}[t!]
    \centering
     \begin{subfigure}[b]{\textwidth}
         \centering \includegraphics[width=\textwidth]{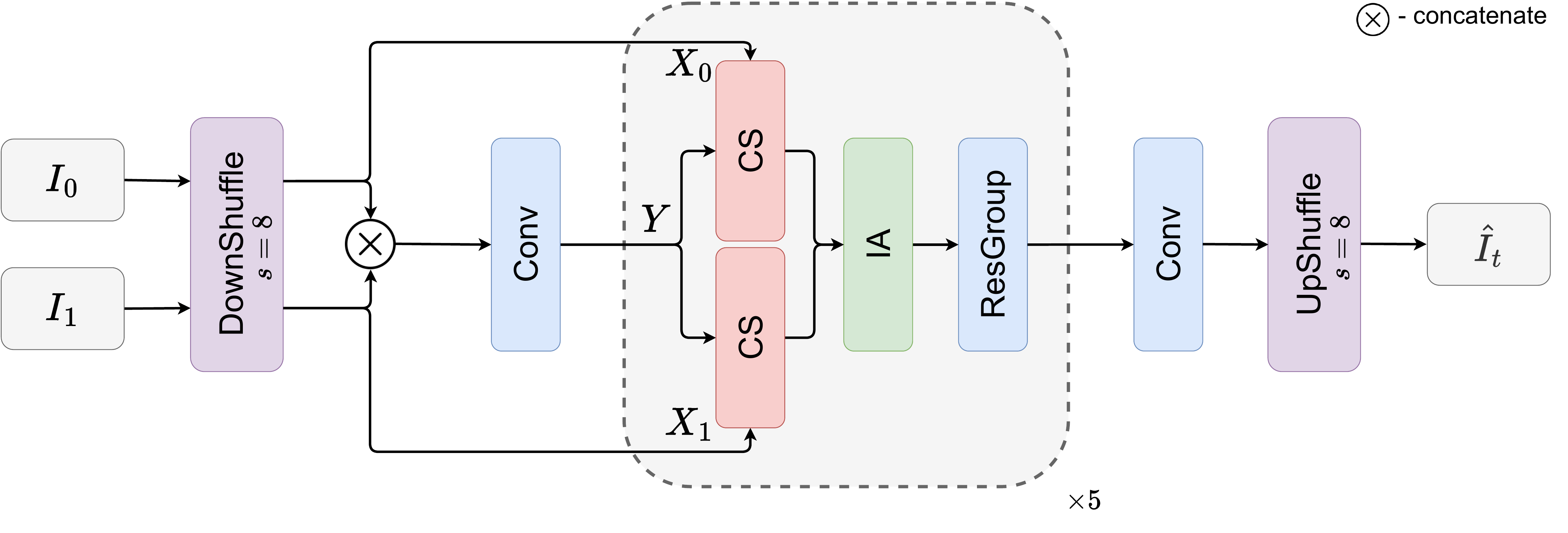}
         \caption{TAIN} \label{fig:overview}
     \end{subfigure}
     \begin{subfigure}[b]{0.712\textwidth}
        \centering 
        \includegraphics[width=\textwidth]{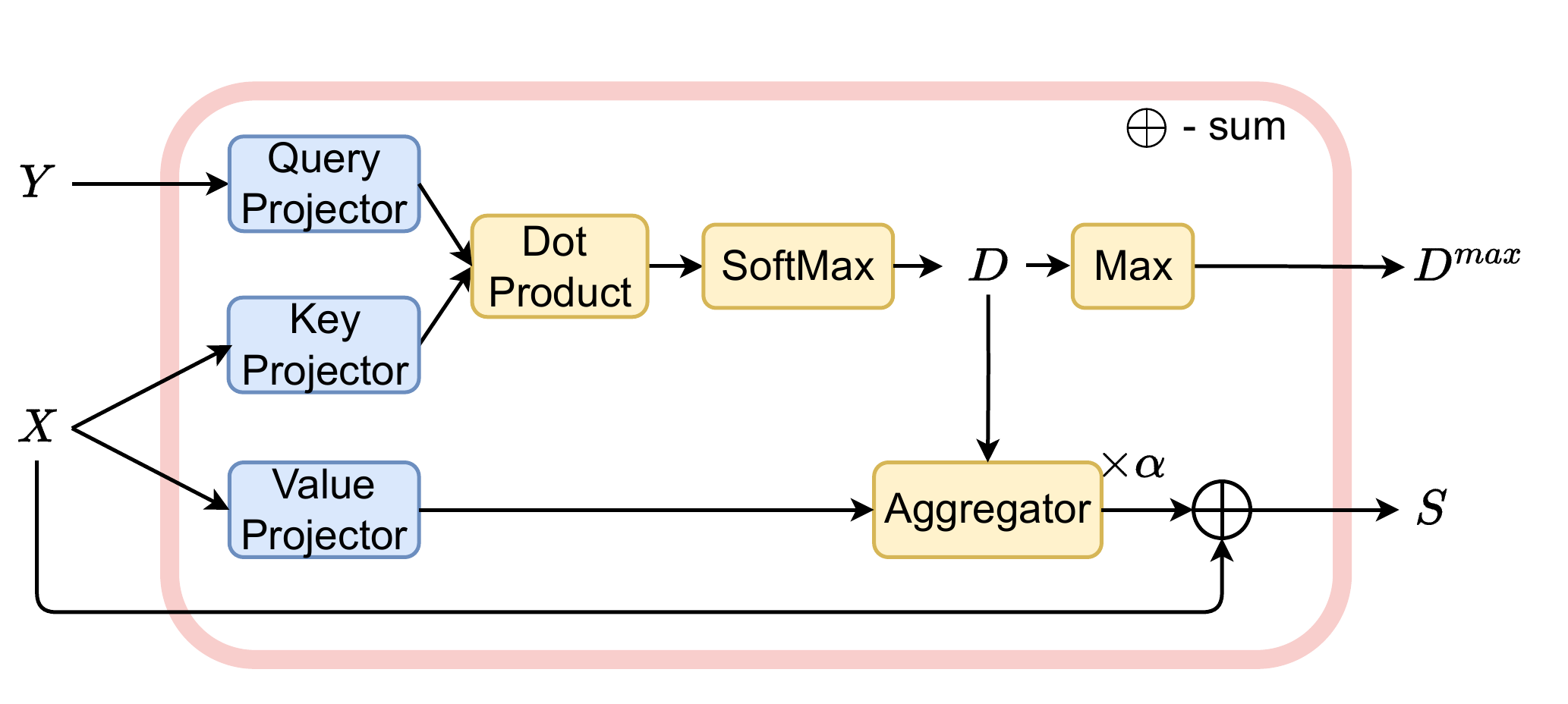} 
         \caption{CS}
         \label{fig:sa}
     \end{subfigure}
     \begin{subfigure}[b]{0.27\textwidth}
         \centering 
         \includegraphics[width=\textwidth]{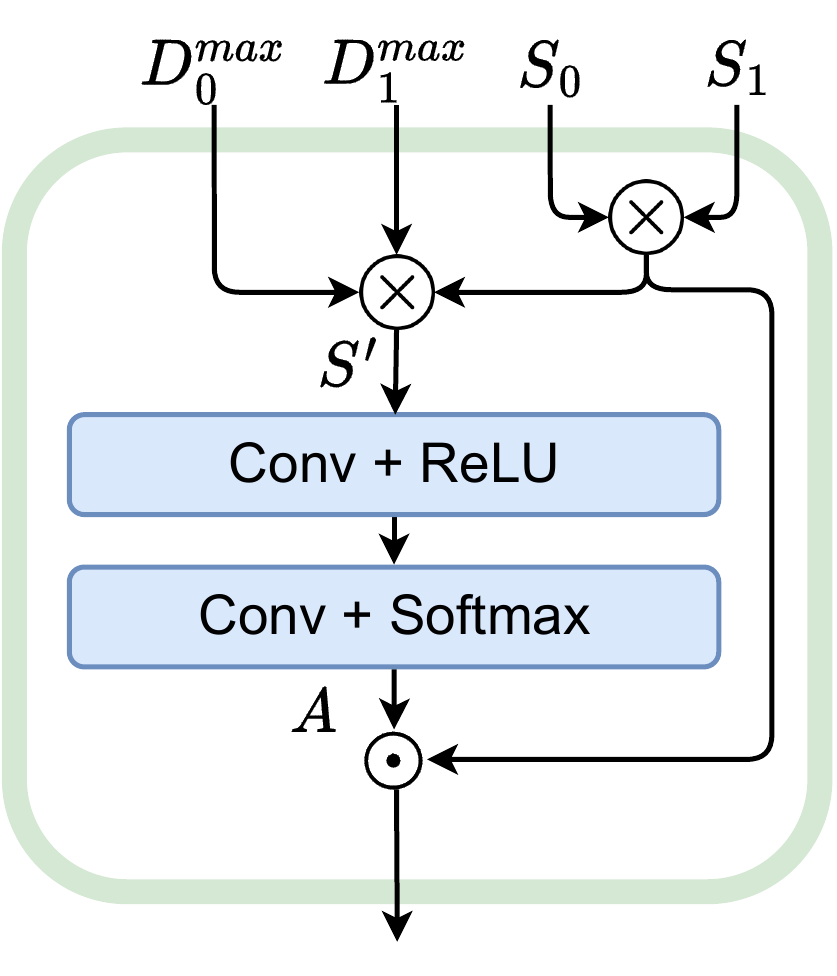} 
         \caption{IA}
         \label{fig:ia}
     \end{subfigure}
    \caption{(a) Overview of the proposed TAIN network for video interpolation. The two consecutive frames $I_0$ and $I_1$ are down-shuffled and concatenated along the channel dimension. They are then processed with five residual groups (ResGroups) with a (b) CS transformer and an (c) IA module before being up-shuffled back to the original resolution to yield the final prediction $\hat{I}_{t}$. The output from each ResGroup is a refinement of the output from the previous block.
    }
    \label{fig:tain}
\end{figure}

\section{Method}

The proposed TAIN method for video interpolation aims to predict frame $I_t \in \mathbb{R}^{h \times w \times 3}$ at time $t=0.5$, given two consecutive images $I_0, I_1\in \mathbb{R}^{h \times w \times 3}$ at times 0 and 1. We do not require any computation of flow, adaptive convolution kernel parameters, or warping, but instead utilize cross-similarity transformer and spatial attention mechanism. Specifically, a novel vision transformer module called the Cross Similarity (CS) module globally aggregates features from input images $I_0$ and $I_1$ that are similar in appearance to those in the current prediction $\hat{I_t}$ of frame $I_t$ (Section \ref{sec:CS}). These aggregated features are then used to refine the prediction $\hat{I_t}$, and the output from each residual group is a new refinement. To account for occlusions of the interpolated features in the aggregated CS features, we propose an Image Attention (IA) module to enable the network to prefer CS features from one frame over those of the other (Section \ref{sec:IA}). See Figure \ref{fig:tain} for an overview of the network. Before describing our network we summarize CAIN~\cite{cain}, on which our work improves.

\subsection{CAIN}
CAIN~\cite{cain} is one of the top performers for video interpolation and does not require estimation of flow, adaptive convolution kernels, or warping. Instead, CAIN utilizes PixelShuffle~\cite{ps} and a channel attention module. PixelShuffle~\cite{ps} rearranges the layout of an image or a feature map without any loss of information. To down-shuffle, activation values are merely rearranged by reducing each of the two spatial dimensions by a factor of $s$ and increasing the channel dimension by a factor of $s^2$. Up-shuffling refers to the inverse operation.
This parameter-free operation allows CAIN to increase the receptive field size of the network's convolutional layers without losing any information.
CAIN first down-shuffles ($s=8$) the two input images and concatenates them along the channel dimension before feeding them to a network with five ResGroups (groups of residual blocks). With the increased number of channels, each of the blocks includes a channel attention module that learns to pay attention to certain channels to gather motion information. 
The size of the features remains $h/8\times w/8\times 192$ throughout CAIN, and the final output map is up-shuffled back to the original resolution of $h \times w \times 3$.

\subsection{Cross Similarity (CS) Module}\label{sec:CS}

All points in the predicted frame $I_t$ appear either in $I_0$ or $I_1$ or in both, except in rare cases where a point appears for a very short time between the two consecutive time frames. These ephemeral apparitions cannot be inferred from $I_0$ or $I_1$ and are ignored here. For the remaining points, we want to find features of $I_0$ or $I_1$ that are similar in appearance to the features in the predicted intermediate frame $\hat{I}_t$, and use them to refine $\hat{I}_t$. We use a transformer to achieve this, where we compare each feature from $\hat{I}_t$ with features from $I_0$ and $I_1$, and use similar features to refine $\hat{I}_t$. 

\begin{figure}[t!]
    \centering
    
    \begin{tabular}{c|ccccc}
       
    \includegraphics[width=0.16\linewidth]{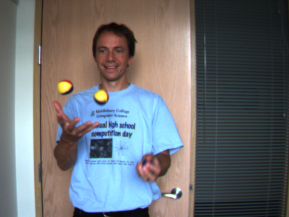} & \includegraphics[width=0.16\linewidth]{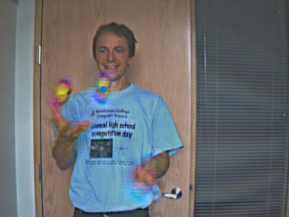} & \includegraphics[width=0.16\linewidth]{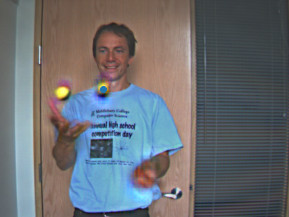} & \includegraphics[width=0.16\linewidth]{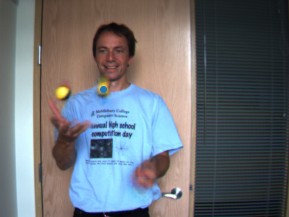} & \includegraphics[width=0.16\linewidth]{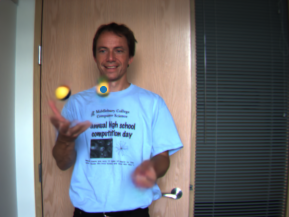} & \includegraphics[width=0.16\linewidth]{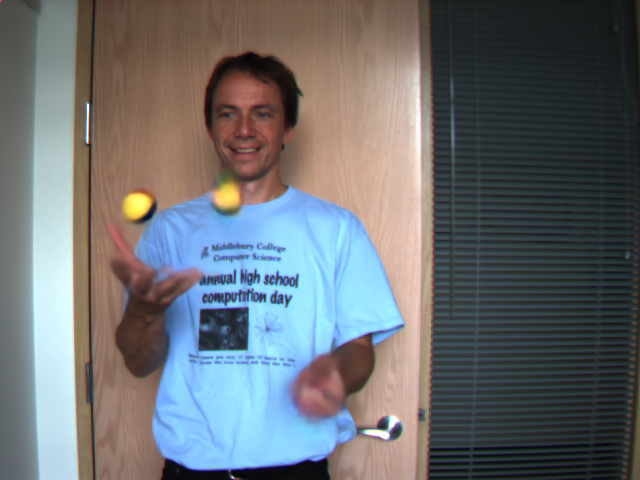} \\
    $I_0$ & $\hat{I_t}$ (r=1)& $\hat{I_t}$ (r=2) & $\hat{I_t}$ (r=3) & $\hat{I_t}$ (r=4) & $\hat{I_t}$ (r=5)\\ 
    
    \includegraphics[width=0.16\linewidth]{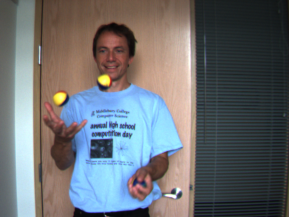} & \includegraphics[width=0.15\linewidth]{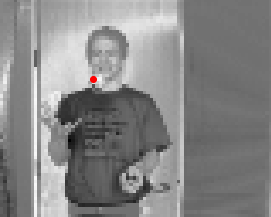} & \includegraphics[width=0.15\linewidth]{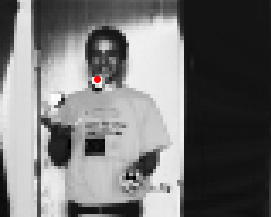} & \includegraphics[width=0.15\linewidth]{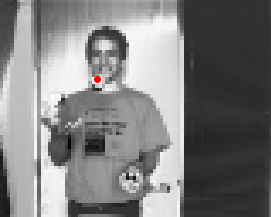} & \includegraphics[width=0.15\linewidth]{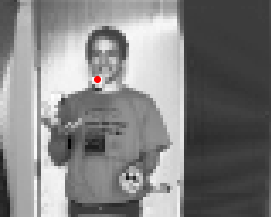} &
    \includegraphics[width=0.16\linewidth]{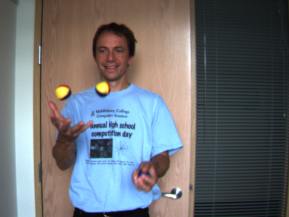} \\
    $I_1$ & $D_1$ (r=1)& $D_1$ (r=2) & $D_1$ (r=3) & $D_1$ (r=4) & $I_t$ \\ 
    \end{tabular}
    \caption{Visualization of intermediate predictions and their cross similarity maps $D_1$ with $I_1$ across all five ResGroups ($r=1\dots5$) using an example from Middlebury~\cite{middlebury} dataset. The first column shows the two input frames, $I_0$ (top) and $I_1$ (bottom). The next four columns show the predicted intermediate frame $\hat{I_t}$ after each of the first four residual blocks with a query point highlighted in blue (top), and its corresponding similarity map $D_1$ from our CS module with the point of highest similarity highlighted in red (bottom). The values in the similarity maps $D_1$ show large scores (closer to white) whenever query and key features are similar in appearance. The last column shows the final output from TAIN (top), which is not used in any CS module, and the ground-truth intermediate frame $I_t$ (bottom) as a reference. }
    \label{fig:CSvis}
\end{figure}

While transformers typically use self-similarity~\cite{vaswani2017attention,nlp_att}, wherein query, key, and value are based on similarities within the same feature array, we extend this notion through the concept of cross-image similarity. Specifically, we use features from the input images $I_0$ or $I_1$ (first column in Figure \ref{fig:CSvis}) as keys and values, and features from the current frame prediction $\hat{I}_t$ (the remaining images in the first row of Figure \ref{fig:CSvis}), as queries. Given a query feature normalized to have unit Euclidean norm (\textit{e.g.}, features for the blue points on the ball in the first row of Figure \ref{fig:CSvis}), our CS module compares it to all the similarly-normalized key features through a dot product, which are then Softmax-normalized to obtain the corresponding similarity maps $D$ (\textit{e.g.}, images in the second row of Figure \ref{fig:CSvis}). The similarity matrix $D$ is used to find the location of the largest similarity score (\textit{e.g.}, red dots in the the second row of Figure \ref{fig:CSvis}), where we retrieve our value features (equation (\ref{eq:sa}) later on), which are projections of the input image features, to yield aggregated input features $S$ based on similarity. These aggregated features are then used to refine the intermediate prediction of $\hat{I}_t$. 

\setlength{\tabcolsep}{1pt} 

\subsubsection{Mathematical Formulation}
Let $X \in \mathbb{R}^{h/s\times w/s \times d}$ denote the (down-shuffled) feature map from one of the input images $I_0, I_1 \in \mathbb{R}^{h\times w\times 3}$ and let $Y \in \mathbb{R}^{h/s\times w/s \times d}$ denote the feature map of the predicted intermediate frame $\hat{I}_t \in \mathbb{R}^{h\times w\times 3}$. Let $M^\mathbf{i}$ represent feature at pixel location $\mathbf{i}$ in a given feature map $M$. Our CS module computes query feature map $Q\in \mathbb{R}^{h/s\times w/s \times d}$ from $Y$. It also computes key feature map $K\in \mathbb{R}^{h/s\times w/s \times d}$ and value feature map $V\in \mathbb{R}^{h/s\times w/s \times d}$ from $X$ as follows for all pixel locations $\mathbf{i}, \mathbf{j}$:
\begin{equation}
    Q^\mathbf{i} = W_{qk}Y^\mathbf{i}, \quad
    K^\mathbf{j} = W_{qk}X^\mathbf{j},\quad
    V^\mathbf{j} = W_vX^\mathbf{j} \;.
\end{equation}
The matrices $W_{qk}, W_v \in \mathbb{R}^{d\times d}$ are learnable. Note that we use cross-attention where the
query and key features are the same projections (shared weights $W_{qk}$) of different features. This is different from the self-attention transformers used in existing work~\cite{Lu_2022_CVPR,Shi_2022_CVPR}, where the query, key, and value features are different projections (no shared weights) of the same features. 

Each query feature $Q^\mathbf{i}$ is compared with all the key features $K^\mathbf{j}$ to compute a similarity matrix $D^\mathbf{i}\in \mathbb{R}^{h/s\times w/s}$ that captures their similarity:
\begin{equation}
    D^\mathbf{i}  = sim(Q^\mathbf{i}, K^\mathbf{j}) = \dfrac{\exp({Q^\mathbf{i}}^TK^\mathbf{j}/\sqrt{d})}{\sum_\mathbf{j}{\exp({Q^\mathbf{i}}^TK^\mathbf{j}/\sqrt{d})}} \;.
\end{equation}

Using this similarity matrix $D^\mathbf{i}$, the CS module finally computes the aggregated similarity feature $S^\mathbf{i}\in \mathbb{R}^{h/s \times w/s \times d}$ by taking the value feature $V^{\mathbf{i}_{\max}}$ at location $\mathbf{i}_{\max}$ corresponding to the maximum similarity score for each query feature $Q^\mathbf{i}$ and adding the result to $Y^\mathbf{i}$ 
\begin{equation}
    S^\mathbf{i} = Y^\mathbf{i} + \alpha V^{\mathbf{i}_{\max}} \;, 
    \label{eq:sa}
\end{equation}
where $\alpha$ is a learnable scalar parameter initialized to zero.

\subsection{Image Attention (IA) Module}\label{sec:IA}

\setlength{\tabcolsep}{1pt} 
\begin{figure}[t!]
    \centering
    \begin{tabular}{ccccc}
        \includegraphics[height=0.15\linewidth]{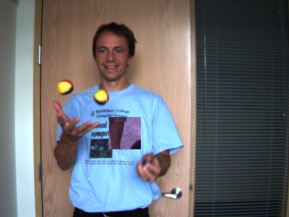} & 
        \includegraphics[height=0.15\linewidth]{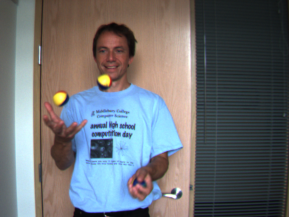} & 
        \includegraphics[height=0.15\linewidth]{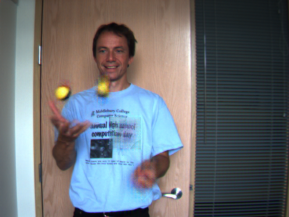} & 
        \includegraphics[height=0.15\linewidth]{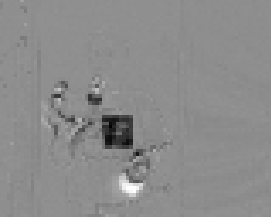} & 
        \includegraphics[height=0.15\linewidth]{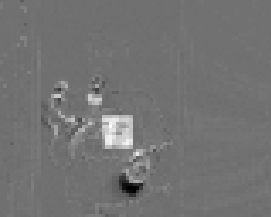} \\
        (a) $I_0$ & (b) $I_1$ & (c) $\hat{I_t}$ & (d) $A_0$ & (e) $A_1$ \\
    \end{tabular}
    \caption{Visualization of Image Attention maps (d) $A_0$ and (e) $A_1$ using an example from Middlebury~\cite{middlebury} (White is large and black is small). For visualization purposes only, we add a square patch to (a) $I_0$ on the person's chest, and show that the IA module assigns a higher weight to the CS features from $I_1$ on the person's chest in (c) $\hat{I_t}$ than to those from $I_0$, which the patch occludes.}
    \label{fig:IAmaps}
\end{figure}

We propose an Image Attention (IA) module based on spatial attention~\cite{Zhang_2018_CVPR} to weigh our two CS features $S_0$ and $S_1$ computed from the input frames as shown above. The IA module enables TAIN to prioritize or suppress features from one input image over those from the other for a given spatial location. This is useful especially at occlusions, where a feature from $I_t$ appears in one input image (likely yielding high similarity scores) but not in the other (likely yielding low similarity scores). This also helps on motion boundaries, where features encode information from both sides of the boundary. As the two sides move relative to each other, the specific mixture of features changes. The IA module compares the two CS feature maps $S_0$ and $S_1$ to construct two IA weight maps $A_0, A_1\in [0, 1]^{h/s \times w/s \times 1}$ where $A_0+A_1 = 1^{h/s \times w/s \times 1}$ as shown below. These weight maps are multiplied with the corresponding CS maps $S_0$ and $S_1$ before they are concatenated and fed to the next ResGroup. See Figure \ref{fig:IAmaps}.

\subsubsection{Mathematical Formulation}

Let $S' \in \mathbb{R}^{h/s \times w/s \times 2(d+1)}$ be the concatenation of $S_0$, $S_1$, and the two maximum similarity maps $D_0$ and $D_1$ along the channel dimension. Our IA module first computes image attention weights $A \in [0,1]^{h/s \times w/s \times 2}$ by applying two $1\times 1$ convolutional layers with ReLU and Softmax on $S'$ as shown in Figure \ref{fig:ia}. More formally, we compute $A$ as:
\begin{equation}
    A = \sigma (W_2 * (\rho (W_1 * S')))
\end{equation}
where $\sigma(\cdot)$ denotes the Softmax function, $\rho(\cdot)$ denotes the ReLU function, and $W_1$ and $W_2$ are the weights of the two $1 \times 1$ convolution layers. This image attention weight map $A$ can be seen as a concatenation of maps $A_0, A_1 \in [0,1]^{h/s \times w/s \times 1}$ where $A_0+A_1 = 1$ which are used to weigh the aggregated similarity features $S_0$ and $S_1$ to obtain the weighted features $\tilde{S}_0$ and $\tilde{S}_1$:
\begin{equation}
    \tilde{S_0} = A_0 \odot S_0 \quad \text{ and } \quad
    \tilde{S_1} = A_1 \odot S_1\;,
\end{equation}
where $\odot$ is the element-wise product.  
Note that this is different from the channel attention module from CAIN~\cite{cain}. The channel attention module computes a $(2d)$-dimensional vector weight over the channel dimension, while the IA module computes a $h/s \times w/s$ weight over the spatial dimension. 

\subsection{TAIN Architecture and Training Details}\label{sec:training}
Figure \ref{fig:overview} shows an overview of the TAIN network for video interpolation. TAIN extends CAIN by applying the proposed CS transformer and IA module after each of the intermediate ResGroups. Each CS transformer obtains query features $Q$ from the features $Y$ from the previous ResGroup, and key and value features from one of the two input image features $X_0$ or $X_1$. We remove the residual connections around each ResGroup so that the output from each ResGroup is a new refinement of the output from the previous ResGroup. Figure \ref{fig:CSvis} shows sample predictions after each of the five ResGroups that the query features are based on. With the IA module, the CS features $S_0$ and $S_1$ are weighted based on their maximum similarity scores from $D_0$ and $D_1$. The two weighted CS features, $\tilde{S}_0$ and $\tilde{S}_1$, are then used to refine the prediction in the next ResGroup.

In order to train the network for cases of occlusions and large motion, we add synthetic moving occlusion patches to the training data. Specifically, we first randomly crop a square patch of size between $21\times 21$ and $61\times 61$ from a different sample in the training set. This cropped patch is pasted onto the input and label images and translated in a linear motion across the frames. 
We apply this occluder patch augmentation to first pre-train on TAIN and then fine-tune on the original training dataset. Following CAIN~\cite{cain}, we also augment the training data with random flips, crops, and color jitter.

Following the literature~\cite{cain,softsplat,sepconv++,vimeo,superslomo,Lee_2020_CVPR}, we train our network using the $L_1$ loss on the difference between the predicted $\hat{I_t}$ and true $I_t$ intermediate frames: $\|\hat{I_t} - I_t\|_1$. As commonly done in the flow literature~\cite{igradient1,igradient2}, we also include an $L_1$ loss on the difference between the gradients of the predicted and true intermediate frames: $\|\nabla\hat{I_t} - \nabla I_t\|_1$.
We use the weighted sum of the two losses as our final training loss:
$\mathcal{L}=\|\hat{I_t} - I_t\|_1 + \gamma\|\nabla\hat{I_t} - \nabla I_t\|_1 $, where $\gamma = 0.1$.

Another common loss used in the literature~\cite{cain,Niklaus_ICCV_2017,softsplat,superslomo,Lee_2020_CVPR} is perceptual loss: $\|\phi \left(\hat{I_t}\right) - \phi \left(I_t\right)\|^2_2$,
where $\phi \left(\cdot\right)$ is a feature from a ImageNet pretrained VGG-19. 
As this loss depends on another network, adding to the computational complexity, we do not use this loss and still show performance improvements.

We implement TAIN in PyTorch~\cite{pytorch} \footnote{Code is available at \href{https://github.com/hannahhalin/TAIN}{https://github.com/hannahhalin/TAIN}.} and train our network with a learning rate of $10^{-4}$ through Adam~\cite{adam} optimizer.

\section{Datasets and Performance Metrics}
As customary~\cite{cain}, we train our model on Vimeo90K~\cite{vimeo}, and evaluate it on four benchmark datasets for video interpolation, \textit{i.e.}, Vimeo90K~\cite{vimeo}, UCF101~\cite{ucf}, SNU-FILM~\cite{cain}, and Middlebury~\cite{middlebury}.
Vimeo90K~\cite{vimeo} consists of 51,312 triplets with a resolution of $256 \times 448$ partitioned into a training set and a testing set.
UCF101~\cite{ucf} contains human action videos of resolution $256 \times 256$. For the evaluation of video interpolation, Liu \textit{et al.}~\cite{ucf_vi} constructed a test set by selecting 379 triplets from UCF101. 
The SNU Frame Interpolation with Large Motion (SNU-FILM)~\cite{cain} dataset contains videos with a wide range of motion sizes for evaluation of video interpolation methods. The dataset is stratified into four settings, Easy, Medium, Hard, and Extreme, based on the temporal gap between the frames. 
Middlebury~\cite{middlebury} includes 12 sequences of images for evaluation. The images are combination of synthetic and real images that are often used as evaluation for video interpolation.

Following the literature, we use Peak Signal-to-Noise Ratio (PSNR) and Structural Similarity Index (SSIM)~\cite{ssim} to measure performance. For the Middlebury dataset~\cite{middlebury}, we use Interpolation Error (IE).

\section{Results}

We first compare TAIN with the current state-of-the-art methods on video interpolation with two frame inputs in Table \ref{tab:perf_stoa_all}. Top panel of Table \ref{tab:perf_stoa_all} lists kernel-based methods, \textit{i.e.}, SepConv~\cite{Niklaus_ICCV_2017} and EDSC~\cite{Niklaus_ICCV_2017}, second panel lists attention-based methods, \textit{i.e.}, CAIN~\cite{cain} and TAIN (ours), third panel lists methods based on both kernels and flow estimations, \textit{i.e.}, AdaCoF~\cite{Lee_2020_CVPR}, MEMC~\cite{MEMC-Net}, and DAIN~\cite{dain}, and bottom panel lists flow-based methods, \textit{i.e.}, TOFlow~\cite{vimeo}, CyclicGen~\cite{cycg}, BMBC~\cite{bmbc}, ABME~\cite{abme}, and VFIformer~\cite{Lu_2022_CVPR}.
Bold values show the highest performance in each panel while underlined values show the highest performance across all panels. TAIN outperforms existing kernel-based methods across all benchmarks. Comparing to the flow-based methods, TAIN outperforms all listed methods except for ABME and VFIformer. However, as shown in Figure \ref{fig:inference_time}, the inference times of AMBE and VFIformer are about 4 and 15 times longer, respectively, than those of TAIN. DAIN and BMBC take around 5 times longer than TAIN to inference while performing comparably to TAIN.

\setlength{\tabcolsep}{2.8pt}
\begin{table}[t!]
    \centering
    \caption{Comparison to the existing methods across Vimeo90k, UCF101, SNU, and Middlebury (M.B.) datasets. Top panel shows kernel-based methods (K), second panel shows attention-based methods (Att.), third panel shows methods based on both kernels and flow (K+F), and bottom panel shows flow-based methods. Higher is better for PSNR and SSIM, and lower is better for IE. Bold values show the best performance in each panel, and underlined values show the best performance across all panels. TAIN (ours) outperforms existing methods based on kernel and attention, and performs comparably to those based on flow on Vimeo90k, UCF101, and SNU datasets. 
    }
    \label{tab:perf_stoa_all}
       \begin{tabular}{c|c|cc|cc|cc|cc|cc}
        \toprule
        & \multirow{2}{*}{Method} & \multicolumn{2}{c|}{Vimeo90k}&\multicolumn{2}{c|}{UCF101}& \multicolumn{2}{c|}{SNU-easy}& 
        \multicolumn{2}{c|}{SNU-extreme} & \multicolumn{1}{c}{M.B.} \\
        & &  PSNR & SSIM & PSNR & SSIM  & PSNR & SSIM  & PSNR & SSIM  & IE \\ 
        \midrule
        \multirow{2}{*}{\rotatebox{90}{\usebox3}} 
        & SepConv
         & 33.79 & 97.02 & 34.78 & 96.69 & 39.41 & 99.00 
        & 24.31 & \textbf{84.48} 
        & 2.27 \\
        & EDSC & \textbf{34.84} & \textbf{97.47} & \textbf{35.13} & \textbf{96.84} & \textbf{40.01} & \textbf{99.04}  
        & \textbf{24.39}  & 84.26 & \textbf{2.02}\\
        \midrule
        \multirow{2}{*}{\rotatebox{90}{\usebox2}} & CAIN 
         & 34.65 & 97.29 & 34.91 & 96.88 & 
        39.89 & 99.00 & 24.78 & 85.07 & \textbf{2.28} \\  
        & TAIN (Ours) & \textbf{35.02} & \textbf{97.51}  & \textbf{35.21} & \textbf{96.92} & \underline{\textbf{40.21}} & \textbf{99.05}&  \textbf{24.80} & \textbf{85.25} & 2.35\\
        \midrule
        \multirow{3}{*}{\rotatebox{90}{\usebox4}} 
        & AdaCoF
        &  34.47 & 97.30 &34.90 & 96.80 & \textbf{39.80} & 99.00 
        & 24.31 & 84.39 & 2.24\\
        & MEMC & 34.29 & 97.39 & 34.96 & 96.82 & - &- & -& -& 2.12\\
        & DAIN
        & \textbf{34.71} & \textbf{97.56} & \textbf{34.99} &\textbf{96.83} & 39.73 & \textbf{99.02}& \textbf{25.09} & \textbf{85.84} &\textbf{2.04}\\ 
        \midrule
        \multirow{5}{*}{\rotatebox{90}{\usebox1}} 
        & TOFlow 
        & 33.73 & 96.82 & 34.58 & 96.67 & 39.08 & 98.90 
        & 23.39 & 83.10  
        & 2.15\\
        & CyclicGen
        & 32.09 & 94.90 & 35.11 & 96.84 & 37.72 & 98.40 & 22.70 & 80.83 & - \\
        & BMBC
         &  35.01 & 97.64 & 35.15 & 96.89& 39.90 & 99.03 & 23.92& 84.33 & 2.04 \\ 
        & ABME  & 36.18 & 98.05 & 35.38 & 96.98 & 39.59 & 99.01 & 25.42 & 86.39 & 2.01\\
        
        & VFIformer & \underline{\textbf{36.50}} & \underline{\textbf{98.16}} & \underline{\textbf{35.43}} & \underline{\textbf{97.00}} & \textbf{40.13} & \underline{\textbf{99.07}} & \underline{\textbf{25.43}} & \underline{\textbf{86.43}} & \underline{\textbf{1.82}}\\
        \bottomrule
    \end{tabular}
\end{table}

\setlength{\tabcolsep}{1pt}
\begin{figure}[t!]
    \centering
    \begin{tabular}{cccc}
    \toprule
    \includegraphics[width=0.245\textwidth]{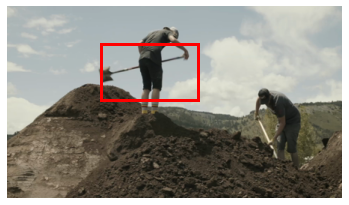} & 
    \includegraphics[width=0.245\textwidth]{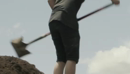}  &
    \includegraphics[width=0.245\textwidth]{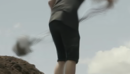}   & 
    \includegraphics[width=0.245\textwidth]{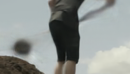} \\
    Input $I_0$ & Label $I_t$ & EDSC & CAIN\\
    \includegraphics[width=0.245\textwidth]{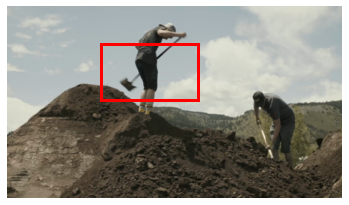} &
    \includegraphics[width=0.245\textwidth]{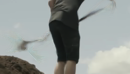} &
    \includegraphics[width=0.245\textwidth]{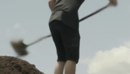} &
    \includegraphics[width=0.245\textwidth]{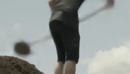} \\ 
    Input $I_1$ & DAIN & VFI & Ours \\ \midrule
    
    \includegraphics[width=0.245\textwidth]{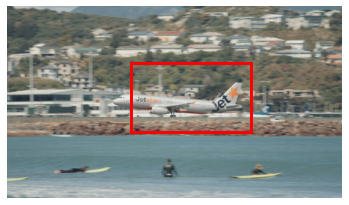} &
    \includegraphics[width=0.245\textwidth]{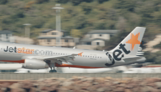} &
    \includegraphics[width=0.245\textwidth]{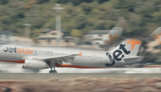} &
    \includegraphics[width=0.245\textwidth]{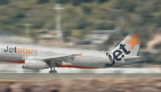} \\
    Input $I_0$ & Label $I_t$ & EDSC & CAIN \\
    \includegraphics[width=0.245\textwidth]{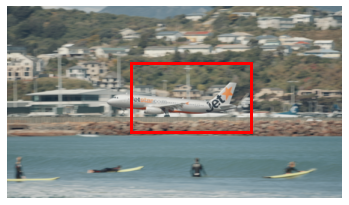} &
    \includegraphics[width=0.245\textwidth]{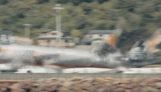} &
    \includegraphics[width=0.245\textwidth]{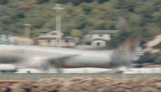}&
    \includegraphics[width=0.245\textwidth]{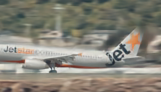}  \\ 
    Input $I_1$ & DAIN & VFI & Ours \\\midrule
    \includegraphics[width=0.234\textwidth]{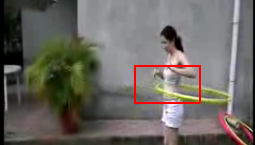} &
    \includegraphics[width=0.245\textwidth]{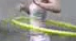} &
    \includegraphics[width=0.245\textwidth]{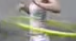} &
    \includegraphics[width=0.245\textwidth]{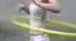} \\
    Input $I_0$ & Label $I_t$ & EDSC & CAIN \\
    \includegraphics[width=0.234\textwidth]{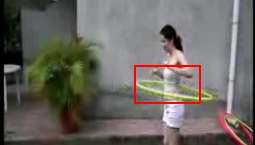} &
    \includegraphics[width=0.245\textwidth]{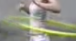} &
    \includegraphics[width=0.245\textwidth]{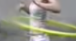}&
    \includegraphics[width=0.245\textwidth]{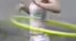}  \\ 
    Input $I_1$ & DAIN & VFI & Ours \\
    \bottomrule
    \end{tabular}
    \caption{Visualization of our proposed method and its comparison to the current state-of-the-art methods~\cite{cain,Niklaus_ICCV_2017,dain,Lu_2022_CVPR} on examples from Vimeo90k and UCF101. }
    \label{fig:vis_stoa}
\end{figure}

Figure \ref{fig:vis_stoa} visualizes examples of predictions from TAIN and the best performing methods in each panel of Table \ref{tab:perf_stoa_all}. Compared with kernel- and attention-based methods, TAIN is able to find the correct location of the moving object, \textit{e.g.} shovel and hula hoop in red box, while keeping the fine details, \textit{e.g.} shaft of the shovel, letters on the plane, and the hula hoop.

\subsubsection{Inference Time}
Figure \ref{fig:inference_time} compares the inference time of TAIN and other state-of-the-art methods listed in Table \ref{tab:perf_stoa_all}. To measure time, we create two random images of size $256\times256$, the same size as those of UCF101 dataset, and evaluate it 300 times using a P100 GPU and report their average and standard deviation. 
As mentioned above, TAIN achieves competitive performance as the existing flow-based methods while taking a fraction of time for inference.

\setlength{\tabcolsep}{3pt}
\begin{figure}[t!]
    \centering
    \begin{tabular}{cccccc}
    \toprule
    \multicolumn{3}{c}{(a) $\hat{I_t}$} & \multicolumn{3}{c}{(b) $I_1$} \\
    \multicolumn{6}{c}{
    \includegraphics[width = 0.93\textwidth]{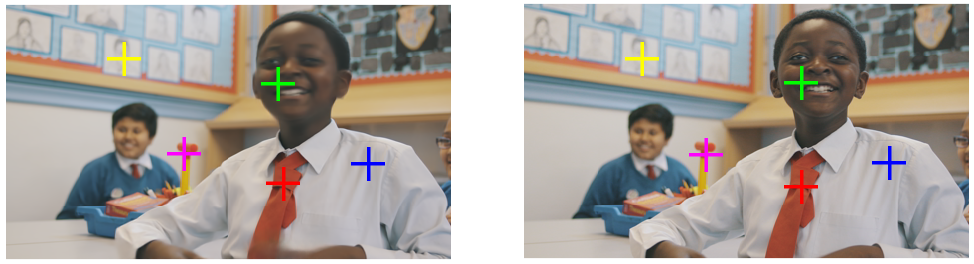}} \\ \midrule
    \raisebox{8.3mm}{\begin{tabular}{c}Query $Q^i$\end{tabular}} & \includegraphics[width = 0.145\textwidth]{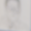} & \includegraphics[width = 0.145\textwidth]{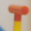} & \includegraphics[width = 0.145\textwidth]{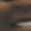} & \includegraphics[width = 0.145\textwidth]{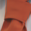} & \includegraphics[width = 0.145\textwidth]{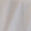}  \\ 
    
    \raisebox{8.3mm}{\begin{tabular}{c}
    Key at \\ 
    $\max{D_1^i}$\end{tabular}} & \includegraphics[width = 0.145\textwidth]{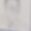} & \includegraphics[width = 0.145\textwidth]{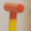} & \includegraphics[width = 0.145\textwidth]{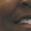} & \includegraphics[width = 0.145\textwidth]{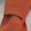} & \includegraphics[width = 0.145\textwidth]{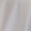}  \\
    
    \raisebox{7.3mm}{\begin{tabular}{c}Similarity \\ Map $D_1^i$\end{tabular}} & \includegraphics[width = 0.145\textwidth]{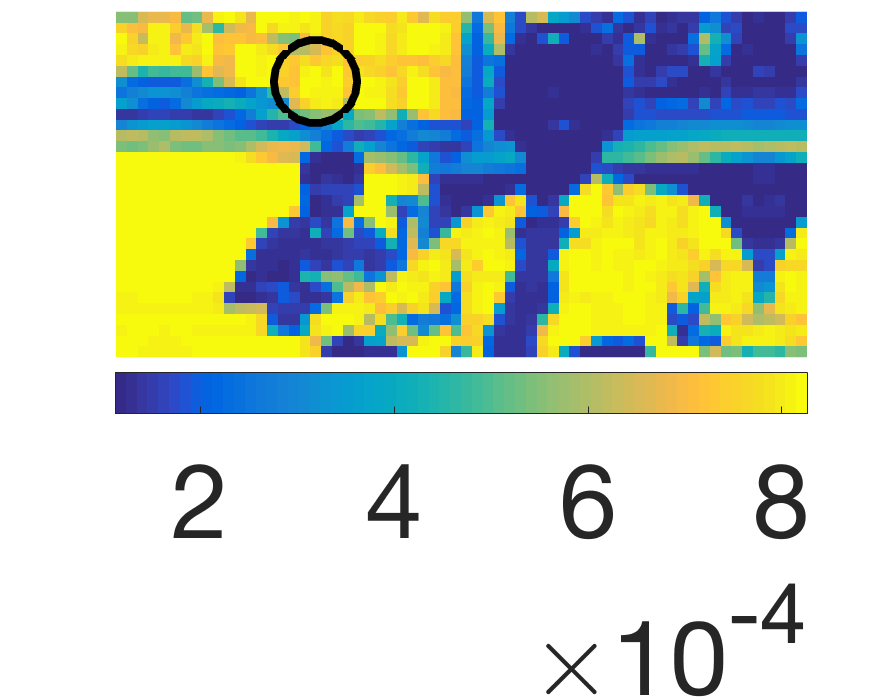} & \includegraphics[width = 0.145\textwidth]{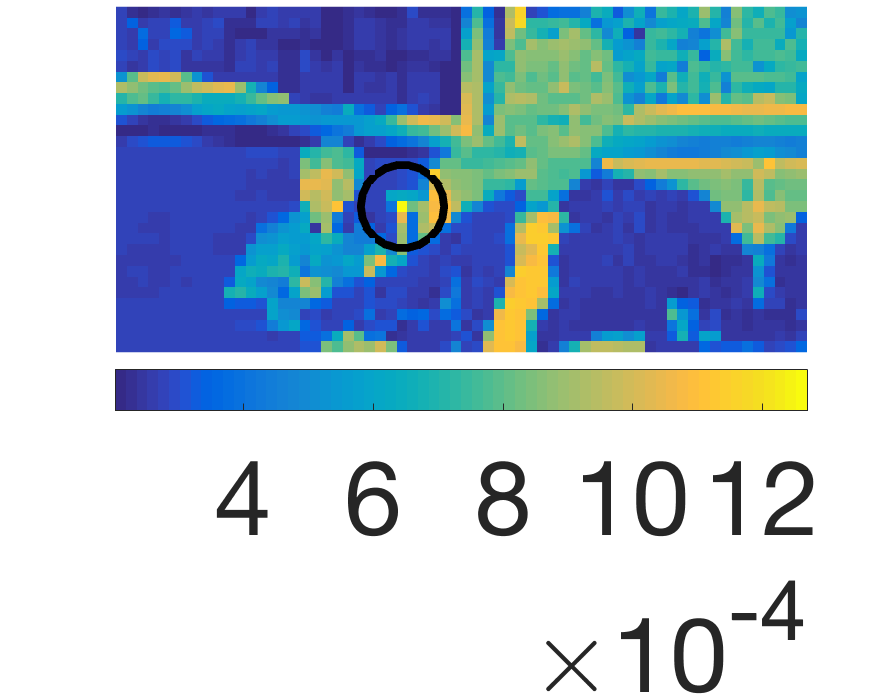} & \includegraphics[width = 0.145\textwidth]{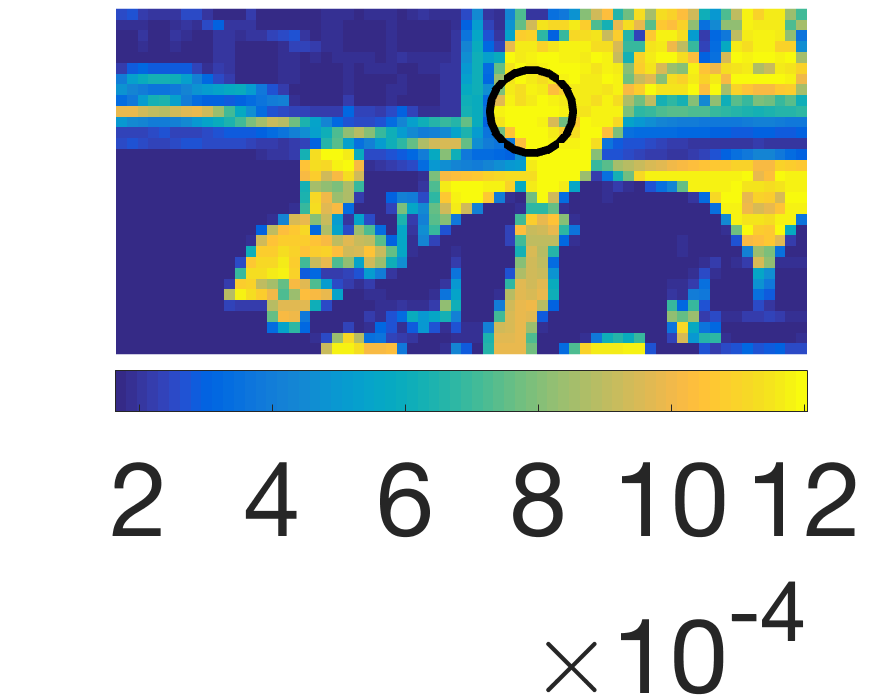} & \includegraphics[width = 0.145\textwidth]{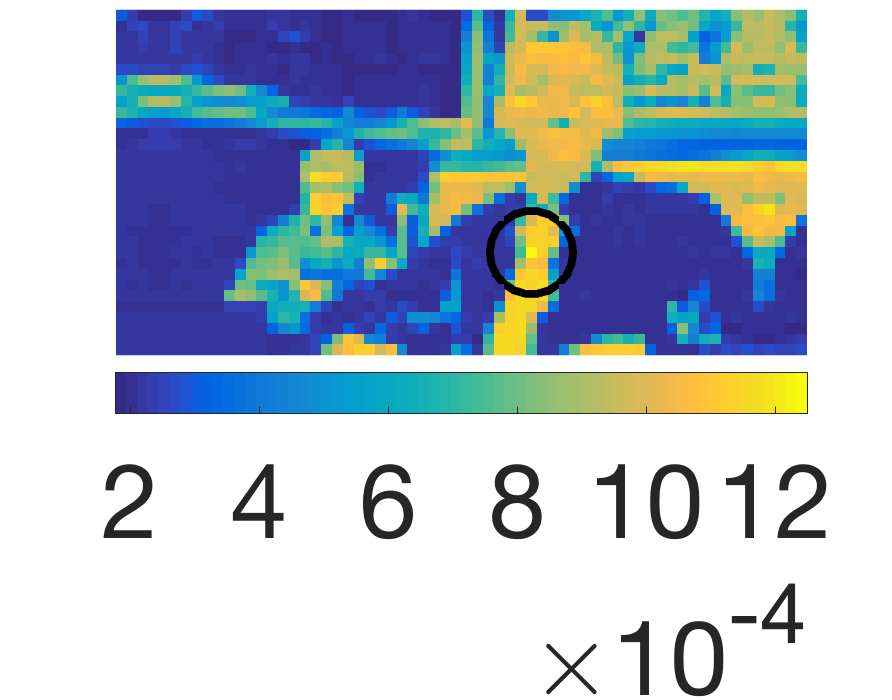} & \includegraphics[width = 0.145\textwidth]{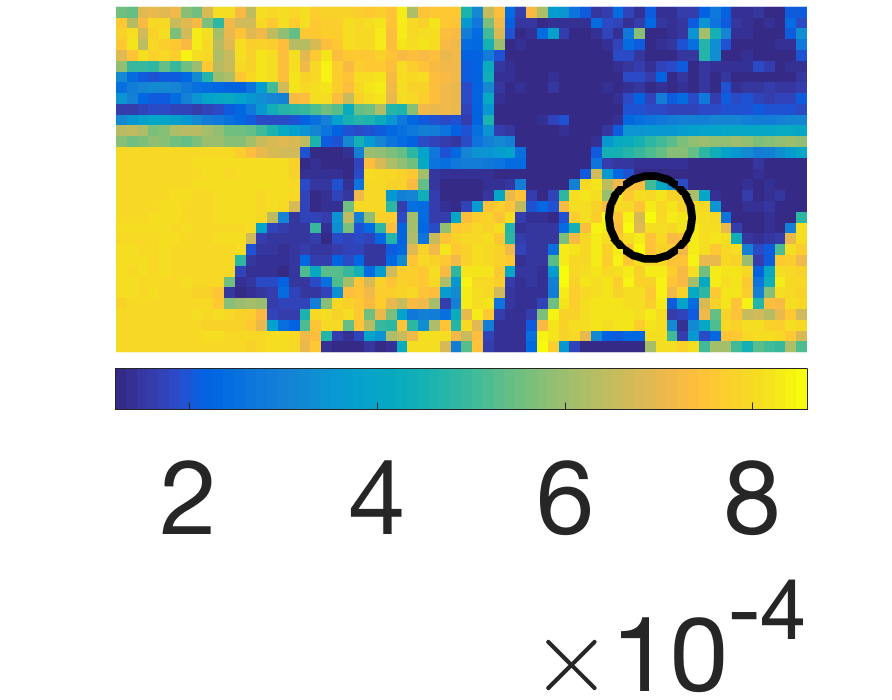}  \\
    \bottomrule
    \end{tabular}
    \caption{Visualization of patches with the highest similarity scores from $D$ in our proposed CS transformer. Top panel shows an example of (a) $\hat{I_t}$ and (b) $I_1$ from the test set of Vimeo90k~\cite{vimeo} dataset. Example query features in $\hat{I_t}$ are shown with `+' mark, and their key feature with the highest similarity score are shown in $I_1$ in corresponding colors. Bottom panel shows $31\times 31$ patches extracted from the query points (`+' in $\hat{I_t}$), their corresponding key patches with the highest similarity score (`+' in $I_1$), and their corresponding similarity map $D_1^i$ with the highest score circled. Our CS module successfully aggregates similar appearance features when refining the interpolation prediction $\hat{I_t}$. 
    }
    \label{fig:att_vis}
\end{figure}

\setlength{\tabcolsep}{3.3pt}
\begin{table}[h!]
\begin{center}
    \caption{Performance changes with our proposed modules:  
    \textbf{IA} - Image Attention module; \textbf{CS} - Cross Similarity transformer; \textbf{\#RG} - Number of ResGroups. Using all the components yields the best overall performance.
    While increasing the number of ResGroups yields consistently higher performance, the performance plateaus after 5 ResGroups, which we use for TAIN (top row).}
    \label{tab:perf_components}
    \begin{tabular}[width=\textwidth]{ccc||cc|cc|cc|cc|c}
    \toprule
    \multirow{2}{*}{IA}
    & \multirow{2}{*}{CS}
    & \multirow{2}{*}{\#RG} & \multicolumn{2}{c|}{Vimeo90k} & \multicolumn{2}{c|}{UCF101}& \multicolumn{2}{c|}{SNU-easy}& 
    \multicolumn{2}{c|}{SNU-extreme} & M.B.\\
    & & & PSNR & SSIM & PSNR & SSIM  & PSNR & SSIM  & PSNR & SSIM & IE\\ \midrule
    \checkmark & \checkmark & 5 & 35.02 & 97.51  & 35.21 & \textbf{96.92} & \textbf{40.21} & \textbf{99.05} & 24.80 & 85.25 & 2.35 \\ \midrule
    
    \checkmark & \checkmark & 4 & 34.82 & 97.38 & \textbf{35.22} & \textbf{96.92} & 40.20& \textbf{99.05} & 24.76 & 84.97 & 2.39 \\
    \checkmark & \checkmark & 6 & \textbf{35.06} & 97.52 & \textbf{35.22} & \textbf{96.92} & 40.20 & \textbf{99.05} & 24.74 & 85.09 & 2.33\\ 
    \checkmark & \checkmark & 7 & \textbf{35.06} & \textbf{97.53} & \textbf{35.22} & \textbf{96.92} & \textbf{40.21} & \textbf{99.05} &24.75& 85.09 & \textbf{2.32} \\ \midrule
    
    & \checkmark & 5 & 34.81 & 97.40 & 35.17 & 96.91 & 40.14 & 99.04 & 24.81 & 85.21 & 2.49 \\\midrule
    &  & 5 & 34.76 & 97.38 & 35.05 & 96.88 & 40.00 & 99.02 & \textbf{24.82} &\textbf{85.28} & 2.66 \\
    \bottomrule
    \end{tabular}
    \end{center}
\end{table}

\section{Ablation Study}

\subsubsection{Visualization of the Components of CS Module}
We visualize the components of our proposed CS module using an example from the test set of Vimeo90k~\cite{vimeo} dataset in Figure \ref{fig:att_vis}. The top panel of this figure shows (a) $\hat{I_t}$ and (b) $I_1$, where example query features $Q^i$ are shown with `+' mark in (a) and their corresponding key features with the highest similarity score are shown with `+' mark in (b) with the corresponding colors. Bottom panel shows $31\times31$ patches extracted from the location `+' of each query $Q^i$ and key $K_1^i$ features from (a) $\hat{I_t}$ and (b) $I_1$, respectively. We also include visualization of the corresponding similarity maps $D_1^i$ and highlight their maximum score. As shown, our CS module successfully extracts similar appearance features when refining $\hat{I_t}$.

\subsubsection{Effect of Each Component}
Table \ref{tab:perf_components} shows the performance changes with varying combinations of our proposed components, \textit{i.e.}, Image Attention (IA), Cross Similarity transformers (CS), and the number of ResGroups (RG). 
Comparing the first four rows that list performances with the changing number of ResGroups from 4 to 7, we see that the performance increases with the number of ResGroups. However, it plateaus after 5 ResGroups, which we choose to use for TAIN (top row) for computational efficiency. In addition, each of the two main components, IA and CS module, contributes to the success of our method.

\section{Conclusion}

We propose TAIN, an extension of the CAIN network, for video interpolation. We utilize a novel vision transformer we call Cross Similarity module to aggregate input image features that are similar in appearance to those in the predicted frame to further refine the prediction. To account for occlusions in these aggregated features, we propose a spatial attention module we call Image Attention to suppress any features from occlusions. Combining both these components, TAIN outperforms existing methods that do not require flow estimation on multiple benchmarks. Compared to methods that utilize flow, TAIN performs comparably while taking a fraction of their time for inference. 

\subsubsection{Acknowledgments:} This research is based upon work supported in part by the National Science Foundation under Grant No. 1909821 and by an Amazon AWS cloud computing award. Any opinions, findings, and conclusions or recommendations expressed in this material are those of the authors and do not necessarily reflect the views of the National Science Foundation.

\bibliographystyle{accv22/splncs}
\bibliography{bib}

\end{document}